# Stretchable and High-Precision Optical Tactile Sensor for Trajectory Tracking of Parallel Mechanisms

Yiding Nie, Dongliang Fan*, Jiatai Huang, Chunyu Liu, and Jian S. Dai*, *Fellow, IEEE*

*Abstract—* Stretchable sensors indicate promising prospects for soft robotics, medical devices, and human-machine interactions due to the high compliance of soft materials. Discrete sensing strategies, including sensor arrays and distributed sensors, are broadly involved in tactile sensors across versatile applications. However, it remains a challenge to achieve high spatial resolution with self-decoupled capacity and insensitivity to other off-axis stimuli for stretchable tactile sensors. Herein, we develop a stretchable tactile sensor based on the proposed continuous spectral-filtering principle, allowing superhigh resolution for applied stimuli. This proposed sensor enables a high-linear spatial response ($R^2 > 0.996$) even during stretching and bending, and high continuous spatial (7 μm) and force (5 mN) resolutions with design scalability and interaction robustness to survive piercing and cutting. We further demonstrate the sensors' performance by integrating them into a planar parallel mechanism for precise trajectory tracking (rotational resolution: 0.02°) in real time.

## I. Introduction

Stretchable sensors, enabling intelligent and versatile responses to surroundings, have been employed for various soft machines, which promise great potential for soft robotics [1-4], wearable devices [5-7], and human-machine interactions [8][9]. Most existing tactile sensors have demonstrated their sensing capabilities for different types of mechanical stimuli's identification [10-12] based on various sensing principles, including resistive [13-15], capacitive [16-18], magnetic [19][20], and optical principles [21-24]. To achieve quantitative perception, force-position decoupling is a necessary trait of tactile sensors. These existing flexible sensors generally possess lower spatial resolution due to the nonlinear response of soft materials compared with rigid ones. Further, resisting the off-axis mechanical stimuli during precise sensing remains a challenge. Discrete sensing technologies, such as sensor arrays and distributed sensing elements, are the most common approaches for quantitative tactile stimuli identification [5][10][11][14], akin to mechanoreceptors in human fingertips with a typical spatial resolution of 1 mm [25]. Corresponding structures, i.e., involving a grid of sensors or employing distributed sensing elements, have usually been adopted for spatially resolved measurements [21][26-28]. Their spatial resolutions highly depend on the density and distribution of the sensor elements, which requires a complex signal processing system with sophisticated signal traces. Thus, to improve spatial resolution, artificial intelligence-enabled super-resolution technologies have been developed to break the limit of the sensor arrays' physical density [19-29]. This discrete sensing technology relies on a sophisticated layout of sensing elements or machine learning and incurs an inconsistent and requires significant compute power over the measuring range.

Optical fiber sensors seem a viable solution to precise tactile sensing due to various intrinsic properties of light, i.e., intensity, frequency, phase, wavelength spectrum, and polarization. Distributed silica-based optical fiber sensors have already exhibited great sensing capacities for continuously measuring parameter variations and separately detecting multiple stimuli, e.g., strain, pressure, and temperature [30][31]. These optical fibers are inextensible and not applicable to soft devices. In contrast to the above sensors, optical waveguide sensors made of stretchable materials relying on the frustrated total internal reflection for optical measurement have been recently developed to perceive different stimuli. Current stretchable optical waveguide sensors that decoupled different stimuli realized a millimeter scale and discrete spatial resolution by analyzing the frequency of the optical signal output with distributed chromatic blocks and thermochromic layers [5][7][21][22]. To further improve the spatial resolution to the microscale, sensing capabilities need to be further revealed by exploring the properties of light for stretchable optical tactile waveguide sensors. Additionally, a practical approach to resisting the off-axis stimuli for tactile perception with high, continuous spatial resolution and mechanical robustness is highly desired.

Here, we proposed continuous spectral-filtering technology and developed a stretchable self-decoupled tactile sensor for acquiring the spatial location and force magnitude of the applied tactile stimuli with high accuracy and sensitivity. The approach takes advantage of spectrum variation during light transmission in a dyed medium, referred to as a continuous filter, measuring the stimuli's spatial locations according to the optical signal's color (global pattern of the spectrum). Simultaneously, the force magnitude is measured by the brightness (integrated area of the spectrum curve), which is a thoroughly decoupled measurand from color in spectrum analysis. Additionally, this structure provides a high resistance to the off-axis stimuli since its parallel-assembled structure filters out other off-axis light transmission. Furthermore, the continuous spectral

This work was supported by the Key Program of the National Natural Science Foundation of China (grant 52335003), the Postdoctoral Fellowship Program (Grade B) of China Postdoctoral Science Foundation (grant GZB20230286), the Science, Technology and Innovation Commission of Shenzhen Municipality (grant ZDSYS20220527171403009), and the High level of special funds from Southern University of Science and Technology, Shenzhen, China (grant G03034K003), China Postdoctoral Science Foundation (grant 2024M761294). (Corresponding authors: Dongliang Fan; Jian S. Dai).

Yiding Nie, Dongliang Fan, Jiatai Huang, Chunyu Liu, and Jian S. Dai are with the Shenzhen Key Laboratory of Intelligent Robotics and Flexible Manufacturing Systems, Southern University of Science and Technology, Shenzhen 518055, China.

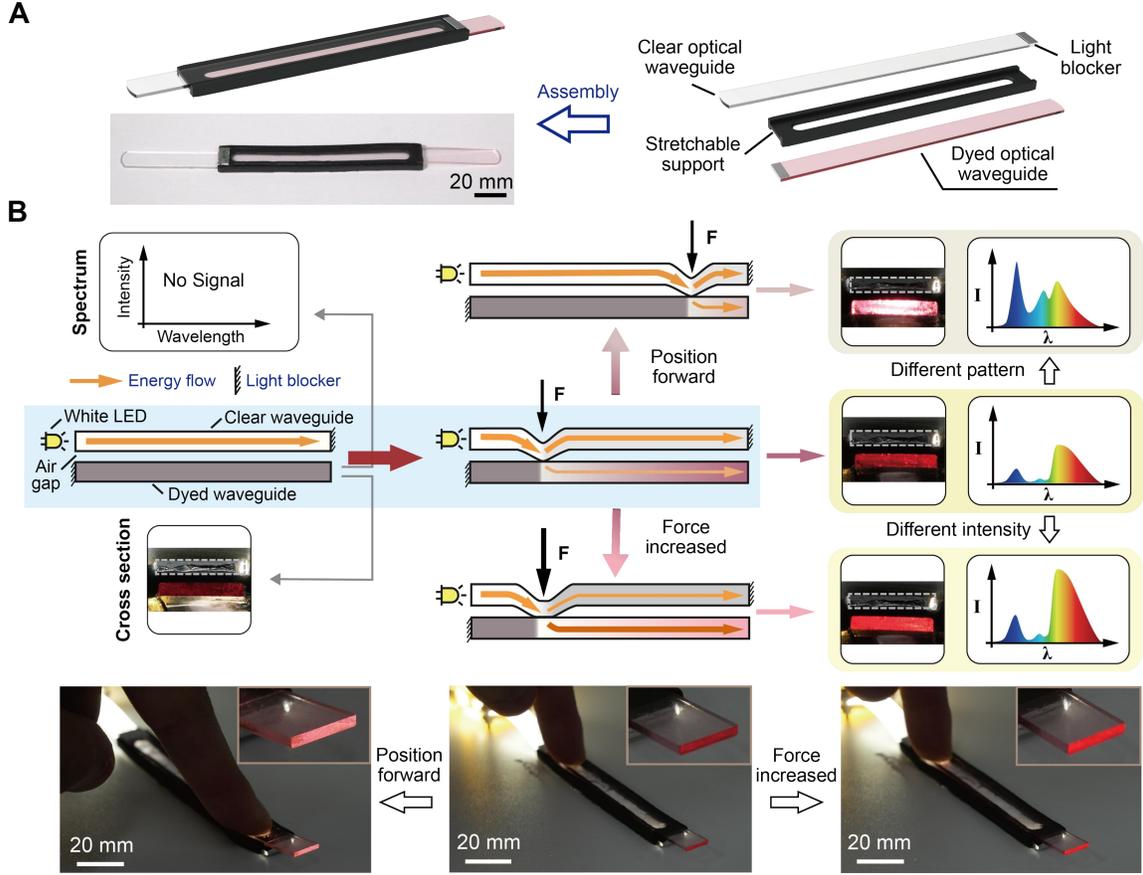

**Figure 1.** Structure and working principle of the optical tactile sensor. (A) Schematic and image of the tactile sensor with main components. (B) Sensor working principle. When an external force is applied to the sensor, the bottom waveguide receives the spectrum information. The position can be calculated by the pattern of the normalized spectrum due to the filtering length, and the decoupled magnitude of the stimuli can be determined by the overall intensity of the spectrum resulting from the change in contact area.

filtering principle and the optical tactile sensor also provide a new approach for optical sensor design and versatile applications. We demonstrated our sensor's applications for proprioception by precisely measuring the trajectory of a planar five-bar linkage and exteroception by human-machine interaction through a mini optical keyboard, a wearable interactive device, and a controller operating in an extreme electromagnetic condition.

## II. Principle and Structure of the Tactile Sensor

### A. Continuous spectral-filtering and contact coupling principles

Absorptive filters with a discrete state are generally employed to modulate the spectral composition of incident light by selectively absorbing specific wavelengths while permitting others to propagate through. Unlike traditional absorptive filters with a dense-dyed and thin medium, we extended the traditional filter and proposed a continuous optical filter with a long-range and slight dye to ensure all wavelengths can transmit through for perceptional purposes. The discrete optical fiber exhibited a step-wise intensity response, while the continuous filter produced a smooth response along the transmitting path. According to the continuous response during the filtering process along the long-range filter, both spatial and spectral information could be conveyed. The intensity response $I_i$ in a dyed medium of an incident light can be determined by:

$$I_i = I_{i0} \cdot e^{-k_i x} \qquad (1)$$

where $I_{i0}$ is the initial intensity of a specific wavelength, $k_i$ is the decay coefficient of the specific wavelength, and $x$ is the filtering length that light paths through in the medium.

Light coupling involves transferring or directing light between different media, a key aspect in optical systems where efficient light transmission between components is critical for ensuring optimal performance. In a parallel-assembled waveguide system, contact coupling is a prevalent approach for realizing light coupling, which has been extensively utilized in optical sensing technologies. When light is incident upon the top optical waveguide, no coupled light is detected in the bottom waveguide in the absence of external force. The applied force could generate a contact area between waveguides, thereby facilitating a new pathway for light to traverse and be transferred to the bottom waveguide. Moreover, the increased force could enlarge the contact area, which results in the increased intensity of the coupled light. Consequently, the applied force can be theoretically derived from the intensity of the coupled light in this waveguide configuration.

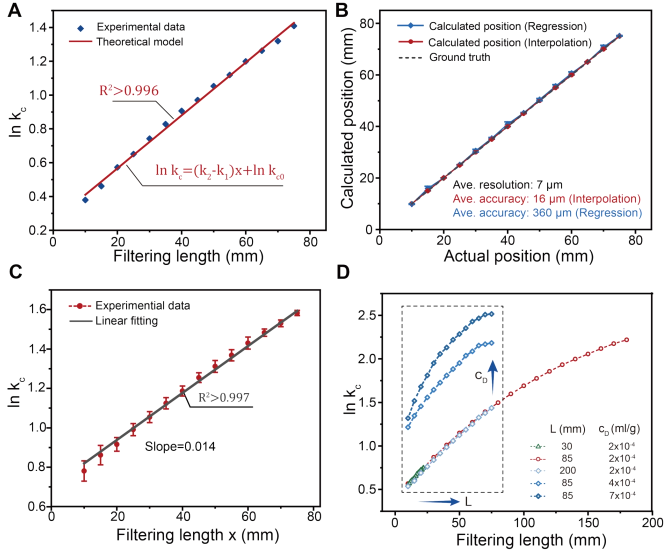

Figure 2. Spatial response of the optical tactile sensor. (A) The linear spatial response to the filtering length of the sensor. (B) Calculated position versus actual position along an 85-mm-long sensor, and the error bar represents the deviation between the calculated value and ground truth. (C) The spatial response to the filtering length of the sensor. The data are presented as mean values ± standard deviation for the number of trials n=3. (D) The effect of dye concentration and waveguide length on the sensor's spatial response.

### B. Structure design and fabrication of the sensor

The optical sensor with a compact structure (3 mm in thickness, 12 mm in width) consist of three layers: a transparent optical waveguide, a hollow support, and a slightly red-dyed optical waveguide. A white LED with a continuous emission spectrum is connected to the transparent waveguide for light transmission. A black-dyed hollow support is employed to generate an air gap between two waveguides. Two aluminum caps are fixed to one end of each waveguide to block environmental disturbance. An RGB sensor rather than a spectrum analyzer was employed to acquire the optical signal to simplify our sensor system and enhance its integrability (Fig. 1A).

Polyurethane optical waveguide fabrication. To fabricate the optical waveguide, part A and part B of Clear Flex 50 (Smooth-On) precursor were first stirred with a volume ratio of 1:2 for 1 minute. Next, the precursor was vacuumed for 5 minutes to degas. Then, the degassed precursor was carefully poured into the mold. The polyurethane precursor was cured at room temperature for 12 hours. The dyed waveguide was prepared by adding red pigment to the precursor and cured.

Silicone support fabrication. The silicone support was prepared by adding Part A, part B of Dragon Skin 30 (Smooth-On) precursors, and part A, part B of Ecoflex 0030 (Smooth-On) precursor with a volume ratio of 2:1. A small amount of black pigment (Silc-Pig, Smooth-On) was added to the precursor, stirred and degassed. After the dyed precursor was injected into a 3D-printed mold, the mold was placed into an oven to cure at 50°C for 2 hours.

Optical sensor assembly. After all components were cleaned by isopropanol, a transparent waveguide, a hollow support, and a red-dyed waveguide were assembled sequentially. Then, two aluminum caps were adhered to the ends of the waveguides by stretchable silicone glue (527S, Jule Tech.). Finally, a 2-watt white LED and an RGB sensor (AS7341, DFROBOT) were installed on the tactile sensor.

### C. Working principle of the sensor

When propagating through a colored medium, polychromatic light experiences differential attenuation of its constituent wavelengths. Consequently, the spectral characteristics of the light vary with the propagation distance in the direction of travel. Therefore, the red-dyed optical waveguide works as a continuous filter for spatial position measurement, and the parallel-assembled structure enables light coupling for force magnitude measurement of tactile stimuli. After an external load is applied to the sensor, the white light from the transparent waveguide flows to the red-dyed optical waveguide via contact coupling, and different wavelengths exhibit different attenuations along transmitting distances to generate corresponding spectrums. During transmission in a red media, for example, blue light (wavelength: 400-500 nm) decays more sharply over red light (wavelength: 600-700 nm). Therefore, as the press position moves away from the light source, the portion of blue light in the output light's spectrum decreases, and the position can be continuously measured by the global pattern of the spectrum. Moreover, the magnitude of the stimuli can also be accurately determined by the illuminance of the optical signal output according to different coupling areas (Fig. 3B). The photographs of the output response of the optical sensor to pressing stimuli with different positions and force magnitudes were also demonstrated in Fig. 3B.

## III. EXPERIMENTAL RESULTS

### A. Characterization of the sensor's spatial response

The continuous spectral-filtering technology achieves a highly linear spatial response with high resolution. Based on the exponential decay of light intensity in dyed mediums during spectrum absorption, the response of the sensor exhibits a high-linear relation ($R^2 > 0.996$) to the position of a pressing (Fig. 2A) in accordance with the theoretical model expectation (Eq.1), which presents the average spatial resolution of 7 μm and accuracy of 16 μm across the scale (Fig. 2B and 2C), comparable with the employed linear stage's resolution. Additionally, the realized spatial resolution has been enhanced by no less than 140 times compared to the previous stretchable ones. The relation between the intensity ratio $I_2 / I_1$ of two incident rays and filtering length $x$ is governed by:

$$\ln \frac{I_2}{I_1} = (k_2 - k_1)x + \ln \frac{I_{20}}{I_{10}} \qquad (2)$$

where $k_1$, $k_2$ are decay coefficients of the two wavelengths in a dyed waveguide and $I_{20} / I_{10}$ is the initial intensity ratio of the light source. The compact structure also provides a high SNR ($> 20$ dB), which is critical to resolution and accuracy. As the theoretical model implies, the sensors can achieve customized scale and sensitivity by selecting appropriate dye concentrations and waveguide lengths, which demonstrates their design flexibility and scalability. Sensors containing the same dye concentration with different lengths (30 mm to 200 mm) share the same spatial response in their overlapped operating range, and the sensitivity can be enlarged by increasing dye concentrations (Fig. 2C) for more precise measurement. Moreover, the deviation of spatial response from the linear theoretical model results from the nonlinear absorbance of the chemical dye due to the variation of absorption rate along the wavelength within the band of measuring channels.

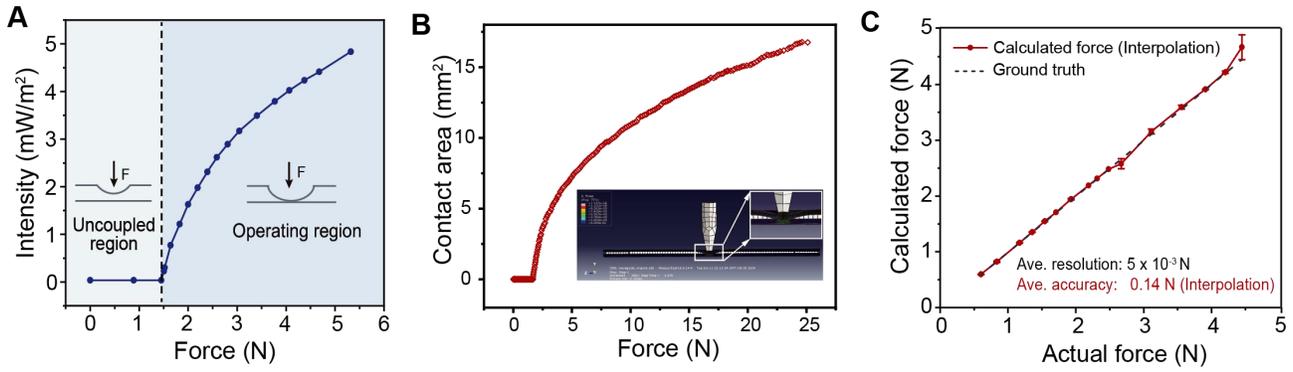

**Figure 3.** Force response of the optical tactile sensor. (A) The response of intensity to applied force. (B) The finite element analysis of the contact area between waveguides with respect to applied force. (C) Calculated force (interpolation) versus actual force along the sensor, and the error bar represents the deviation between the calculated value and ground truth.

### B. Characterization of the sensor's force response

According to the air gap between the two parallel waveguides, an uncoupled region and a threshold force exist before contact coupling is generated. The output signal behaves nonlinearly to the applied force because of the hyper-elasticity of the waveguides in the operating region as shown in Fig. 3A.

Moreover, numerical simulation was also performed to analyze the force response. In the Abaqus modeling and simulation, the behavior of the sensor assembly under a perpendicular force was investigated. The waveguides, each measuring 85 mm by 8 mm by 1 mm, were set as TPU based on their experimental properties, while the supporter and force applier were modeled as rigid bodies due to their negligible deformation. The force applier is positioned centrally on the upper surface of the top waveguide, exerting a downward force along the Y-axis. Boundary conditions included fixing the bottom surface of the lower waveguide and the surrounding surfaces of both waveguides to ensure constrained movement. Meshing details for the waveguides incorporated a global seed size of 0.1, using hexahedron, structured grids with C3D8R element. This precise configuration facilitated a comprehensive analysis of the waveguide's deformation and stress distribution under the applied load, providing valuable insights aligned with the objectives. The result of numerical simulation exhibits a similar tendency with the experiment of force response since the contact area is linearly proportional to light intensity. Despite the nonlinear force response, the sensor still provides 5 mN of average force resolution and 0.14 N of average force accuracy by interpolation (Fig. 3C).

### C. Robustness of optical sensor

Besides high spatial resolution, resisting or predicting off-axis deformation is crucial to precise measurements and practical scenarios. Existing stretchable sensors have the capacity to adapt to different geometries but usually lose their spatial linearity and accuracy when deformed. Our sensor presents slightly different responses to position as the strain increases, which causes the effective dye concentration to decrease (the waveguide's volume increases) due to the Poisson effect of the waveguide material. Still, the sensor retains similar average spatial accuracy even under strain up to 25% (Fig. 4A). Additionally, the response of the sensor subject to various bending (up to 180 degrees) indicates the high decoupling capacity to reject the initial bending deformation due to the existence of the air gap (Fig. 4B).

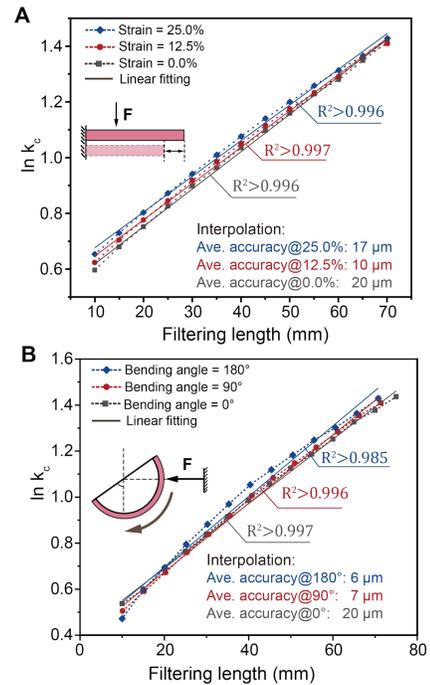

**Figure 4.** Response to off-axis deformations. (A) Sensor responses to filtering length for given initial tensions. (B) Sensor responses to filtering length for given initial curvatures.

The response of the sensor under different stimuli guarantees outstanding adaptability and resilience for scalable measuring ranges, as well as the measuring capacity on curved surfaces to adapt to various applicable scenarios without significant linearity and accuracy loss. The test platform for robustness validation is shown in Fig. 5.

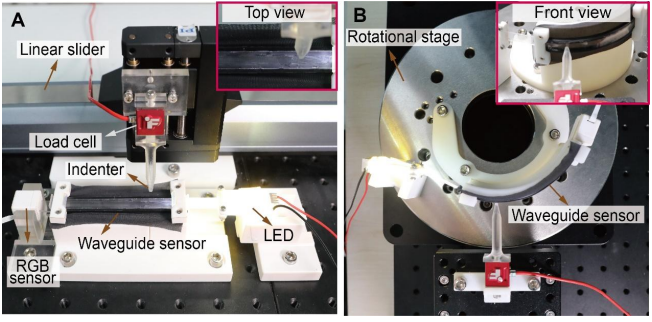
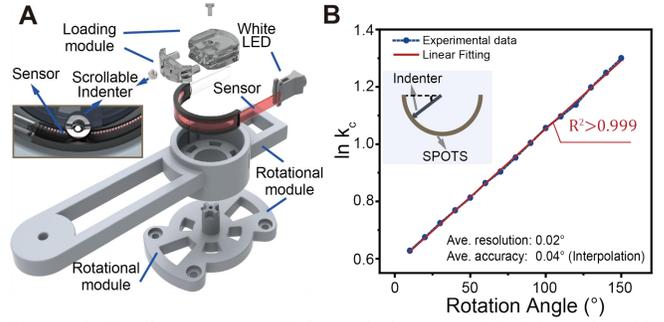

**Figure 5.** Setups for sensor characterization. (A) Test platform for position and force analysis under different strains. The inset shows the top view of the setup. (B) Test platform for position and force analysis under different curvatures. The inset shows the front view of the test platform.

## IV. APPLICATION FOR TRAJECTORY TRACKING OF PARALLEL MECHANISMS

The versatile sensing performance of our continuous spectral-filtering technology enriches the applications of robot-sensor integration. Many examples of soft sensors have been demonstrated for external or internal stimuli recognition on soft robots but seldom presented accurate deformation measurements due to the poor resolution of most existing soft sensors [10][14][32]. Parallel linkages have been employed for various applications, including pick-and-place operations, food packaging, and working as medical operating platforms, owing to their high working speed, prominent operating accuracy, and scalable design [33-35]. Their precise motions rely on high-accuracy encoders for delicate operations and micromanipulations.

**Figure 6.** Bending response of the optical sensor. (A) Schematic with detailed structure of the joint with optical sensor as a soft encoder. (B) The response of sensor to rotational angle.

Here we integrated our optical sensor, working as a soft encoder, into a rotational joint in a highly bending state, as shown in Fig. 6A. A scrollable indenter with a moderate load is installed onto the top surface of the sensor to generate a consistent pressing stimulus. The sensor presented a high linear response ($R^2 > 0.999$) to rotational angle with high spatial resolution (0.02°) and accuracy (0.04°), which is comparable to traditional rigid high-resolution encoders, demonstrating its high accuracy and bending deformation resistance for position analysis, as shown in Fig. 6B.

The continuous high spatial resolution provides a possible routine for delicate manipulation. Therefore, after analyzing the performance of a single soft encoder, we integrated two soft encoders into a planar five-bar mechanism for the trajectory tracking of its terminal (Fig. 7A). This planar five-bar parallel mechanism is a simplified yet highly versatile parallel robotic system, consisting of five interconnected links joined by revolute joints. This mechanism operates in two-dimensional planes and offers two degrees of freedom (2-DOF), enabling intricate trajectory adjustments while maintaining high rigidity and load-bearing capacity. According to the measured angles $\theta_1$ and $\theta_2$ of the two joints by the embedded soft encoders, the terminal T (x, y) of the five-bar linkage could be calculated by the following equations:

$$\theta_1 = \frac{\pi}{2} - \tan^{-1}\frac{y}{x} - \cos^{-1}\left(\frac{\sqrt{x^2+y^2}}{2l}\right) \quad (3)$$

$$\theta_2 = \frac{\pi}{2} - \tan^{-1}\frac{y}{d-x} - \cos^{-1}\left(\frac{\sqrt{(d-x)^2+y^2}}{2l}\right) \quad (4)$$

where $d$ is the distance between two joints and $l$ is the length of the four same bars. Encoders play a critical role in the five-bar mechanisms by providing real-time positional feedback of the driving joints, enabling precise closed-loop control of the two input axes to achieve accurate end-effector trajectories, compensating for mechanical errors. Here, we installed our soft optical sensor, working as the encoder, to record the bending angles of each joint in the parallel mechanism. The output light signal was analyzed by the RGB sensor.

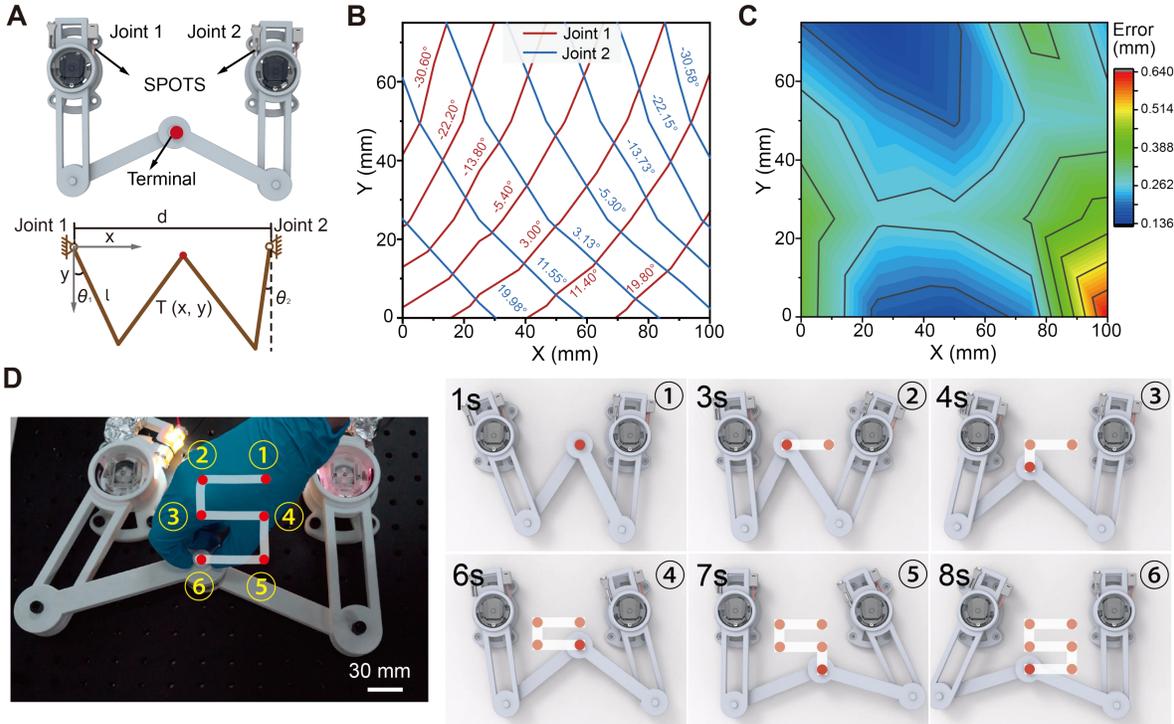

**Figure 7.** Demonstrations of the optical sensor for proprioception. (A) Schematic and mechanism diagram of the planar five-bar linkage with two SPOTS. (B) The output angles of two encoders in relation to the terminal position of the linkage. The red (blue) lines represent positions with the same Joint 1 (Joint 2) output angles. (C) The deviation along the working plane of the five-bar linkage relying on SPOTS. (D) The terminal trajectory "S" of the linkage and its sequential frames of its digital twin in real time.

The terminal position of the mechanism in response to the output angles of two encoders is shown in Fig. 7B. The maximum deviation of the terminal's position is 0.64 mm in its working space (Fig. 7C). Moreover, the systematic error of the five-bar linkage mainly limits its position accuracy. Later, we manually guided the terminal's movement along the trajectory of "SUST", and the digital twin of its motion was also calculated and reconstructed accordingly, relying on the sensors' output signal. The trajectory and even the subtle vibrations of the terminal during movement were precisely replicated, which presenting high spatial resolution and adaptive flexibility for precise trajectory tracking of parallel mechanisms.

## V. CONCLUSION AND FUTURE WORK

In conclusion, we developed continuous spectral-filtering technology and employed the principle into a sensor containing a parallelled-assembled structure for precise tactile perception. This technology enables a 7 μm resolution for continuous high-linearity simultaneous spatial position measurement and 5 mN force resolution based on the linear slider we employed. We also demonstrated its application in trajectory tracking of a parallel mechanism in real time. Furthermore, the sensor's performance paves a promising path toward other interactive interfaces with precision and robustness, such as industrial robots, micromanipulations, and medical operations. Future work will focus on eliminating or reducing the impact of trigger dead zones during external stress measurement processes, enhancing the sensor's stress detection performance to product-grade levels; optimizing the waveguide structure to achieve lower optical energy loss, improving the overall performance of the sensor; and offering design flexibility to facilitate customized fabrication processes and versatile application contexts.